# Pairwise Learning Increases Data Efficiency in Deep Neural Networks

Robert Bain[a*], Marjan Rashidi[b], Elizabeth R. Chrastil[a,b]

[a]Department of Neurobiology & Behavior, University of California, Irvine, Irvine, CA 92697, USA

[b]Department of Cognitive Sciences, University of California, Irvine, Irvine, CA 92697, USA

* Corresponding author: Marjan Rashidi

E-mail addresses: Robert Bain: rbain18@gmail.com, Marjan Rashidi: marjanbain@gmail.com, Elizabeth R. Chrastil: chrastil@uci.edu

**Abstract**

Deep neural networks often fail at small-sample regression, collapsing to low-variance predictions that miss individual differences. Here, we introduce a pairwise comparison framework that reframes regression as aggregated relative rankings. This method improves data efficiency across model architectures and dataset sizes. On the UTKFace dataset (23k images), pairwise learning generates over one billion training comparisons without additional supervision. The paired ranking allows a lightweight ResNet-10 trained from scratch on home hardware to achieve a 3.64-year MAE—surpassing previous models, including those with large-scale pretraining. For classification into categories, pairwise supervision consistently outperforms an otherwise identical 10-way crossentropy model on STL-10 with 500–2000 labeled examples, improving accuracy by up to 5.2%. Overall, pairwise supervision enhances data efficiency and enables strong regression and classification performance when labeled data are limited, particularly in clinical settings.

Keywords: Pairwise Learning, Deep Learning, Small-Sample Learning, Image Classification, Regression

## 1 Introduction

Deep neural networks have transformed supervised learning in machine vision and beyond [1], but they remain notoriously data-hungry and brittle in low-sample regimes [2, 3, 4, 5, 6]. In regression settings, scarce labels often drive models toward degenerate, low-variance predictions that ignore meaningful variation (see Fig. 1 A & B). This behavior is reminiscent of mode collapse in generative adversarial networks (GANs), when a high-variance model converges to a biased solution that fails to reflect the diversity of the underlying distribution [7, 8]. Deep regression instability in reinforcement learning (RL) appears related as well, where deep value and policy networks often exhibit tremendous data inefficiency [9, 10, 11]. Many RL updates reduce to pointwise regression on noisy value targets, and deep networks trained in this manner frequently have large and sudden performance swings while training [12, 9, 13]. Existing responses to this small-data regression problem typically rely on transferring features from large pretrained backbones, hand-crafted data augmentation, or more complex training schemes such as meta-learning, few-shot learning, and semi- or self-supervised methods [14, 15, 16, 17, 18, 19]. While these approaches can mitigate overfitting, they do not fundamentally alter the supervision signal itself. Networks are still asked to predict absolute targets from individual examples, which limits how much information can be extracted from a labeled dataset.

An alternative line of work, dating back to preference learning and learning-to-rank, shows that models can be trained from relative judgments between examples—"which of these two items is larger, better, or more relevant?" [20, 21, 22] —rather than from absolute labels alone. Pairwise and listwise ranking methods, including more recent neural variants, demonstrate that such comparative supervision can be statistically efficient and robust [21, 22, 23]. Architecturally, our use of shared feature extractors for input pairs is related to Siamese and metric-learning networks that learn embeddings from paired or triplet inputs for similarity, verification, or retrieval [24, 25, 26]. Our objective, however, is fundamentally different. Rather than learning a ranking as the final prediction, we use pairwise supervision as an alternative training signal for continuous regression and small-sample classification. Continuous target values are recovered by aggregating comparisons against labeled reference examples, allowing the same set of labels to generate substantially richer supervision without requiring additional annotations or large-scale pretraining. While pairwise ranking has been extensively studied in information retrieval [33, 22] and recommendation systems [34], much less attention has been devoted to using pairwise supervision as a general strategy for improving deep regression and classification under limited-data conditions [35, 36].

In this work we explore a simple pairwise comparison framework that recasts regression as learning from relative differences instead of absolute targets. A single backbone network is trained on labeled pairs to predict which input has the larger underlying label (e.g., which face is older), or for classification of whether the pair of images belong to the same class, using standard cross-entropy loss. At test time, an unlabeled sample is compared against many labeled references, and its value is inferred by aggregating these pairwise outcomes. Unlike learning-to-rank methods (e.g.[33, 34, 22]), where the pairwise comparison itself is the desired output, the comparisons here serve only as an intermediate supervision signal for recovering continuous regression targets. This reframing supplies each example with far richer supervision than typical pointwise regression and circumvents the low-variance collapse common in small-sample settings. We find that this approach yields stable, high-resolution predictions even in datasets with very few ground-truth labels, and that it scales naturally to larger datasets without changing the underlying model or training objective. Beyond regression, we show that the same comparative signal also strengthens supervised classification under severe label scarcity, indicating that pairwise supervision is an upgrade for deep learning in general.

# 2 Methods and Results

To facilitate reproducibility, the code and training scripts used in this work are publicly available at https://github.com/bainro/pairwise_learning. We evaluate the pairwise comparison framework on two settings: (i) age regression on UTKFace and (iii) image classification into object categories on STL-10 under severe label constraints.

## 2.1 Implementation

**Pairwise input construction.** For UTKFace, each training example consists of a pair of images ($x_i$,$x_j$), stacked along the channel dimension and fed into a shared convolutional backbone. The network outputs a single logit indicating whether the age of $x_j$ is greater than that of $x_i$. The target is a binary label $y_{ij} \in \{0,1\}$, and we train with standard cross-entropy.

**Data augmentation and 8× test-time augmentation.** For UTKFace, both ordered comparisons ($x_i$,$x_j$) and ($x_j$,$x_i$), are used as part of the pairwise formulation, with complementary labels. Separately, as image-level augmentation, each face is independently flipped horizontally with probability 0.5. At test time, predictions are averaged across both input orders and the four possible horizontal-flip configurations, yielding eight evaluations per underlying pair (two orders × two flip states for $x_i$ × two flip states for $x_j$). The same image-level augmentation and test-time augmentation are applied to the pointwise baseline to ensure a matched comparison. On the full dataset of roughly $N \approx 23{,}000$ labeled faces, forming all $\binom{N}{2}$ ordered pairs produces on the order of $10^8$ distinct comparisons, with these augmentations, the effective number of pairwise training examples exceeds $10^9$.[1]

STL-10 follows the same pairwise construction as UTKFace including test-time augmentation, except for the labels, which are 1 if the classes are the same in the pair, 0 otherwise.

[1] Exact pair counts depend on the final train/test/ref. We report approximate scales here.

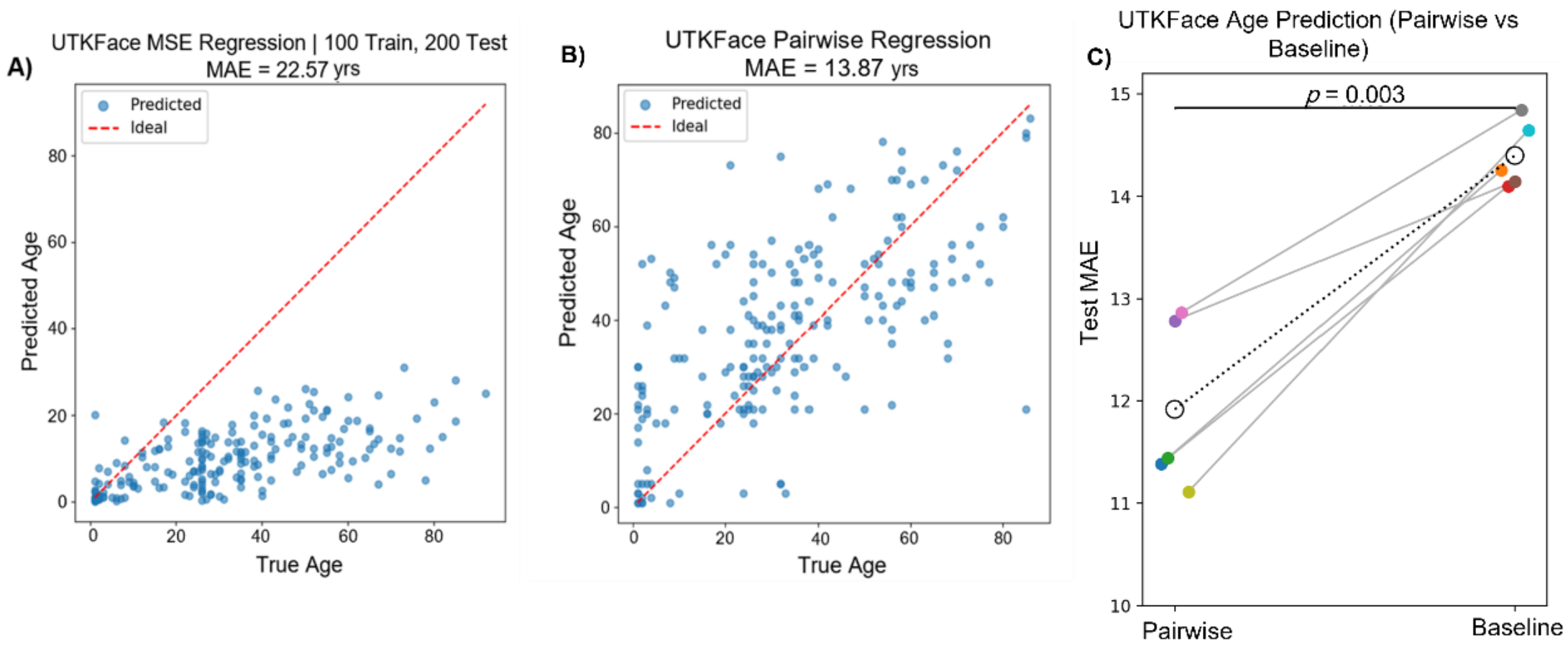

**Figure 1: The Problem: Collapse of pointwise MSE regression. A)** True vs predicted ages from 2D faces for the direct MSE regression baseline. Across subjects, predictions collapse toward a narrow band near the training-set mean, yielding extremely low dynamic range and failing to capture subject-specific variation. The diagonal line denotes ideal predictions, the clustering around a constant value illustrates the characteristic failure of pointwise regression in the small-data regime. This model was trained on only 100 (face, age) pairs to illustrate the difficulty with small sample sizes. **B)** Pairwise learning on UTKface's age regression task with only 100 training data points solves regression collapse (contrast with Fig. 1A). Unlike the pointwise MSE baseline, the pairwise model preserves the full dynamic range of ages predictions. The inferred labels show strong monotonic structure and substantially improved accuracy (test MAE = 13.9 vs. 22.6 for direct MSE regression), demonstrating the benefit of learning from relative comparisons rather than absolute targets in the small-data regime. Test data points were sampled evenly throughout the whole sorted list of subjects to increase the difficulty of the test set and to ensure the best possible test of generality. **C)** Pairwise learning consistently improves regression performance across repeated runs. The figure shows test MAE for pairwise and pointwise baseline models on UTKFace age prediction using 100 training images and 200 test images across five matched random seeds. Results are paired by random seed, such that each pairwise run is compared with the pointwise baseline run using the same seed. Gray lines connect results from the same seed, colored points indicate individual runs, and open circles indicate the mean across seeds. Pairwise learning achieved significantly lower test MAE than the pointwise baseline for UTKFace ($p$ = .003), further demonstrating the data-efficiency afforded by pairwise construction without any additional human annotation.

**Architectures and optimization.** On UTKFace and STL-10, our primary model is a lightweight ResNet-10 backbone made fully convolutional except for the final linear layer, keeping the ratio of effective training comparisons to trainable parameters high. All models are trained from scratch with the Adam optimizer and cross-entropy loss on pairwise labels, using modest learning rates and early stopping on a held-out test set.

**Pairwise learning formulation.** Across all datasets, learning was formulated as a pairwise comparison problem. Training examples consisted of ordered input pairs $(x_i, x_j)$, and models were optimized to predict whether the target value associated with $x_j$ exceeded that of $x_i$. Binary labels were defined accordingly, and optimization was performed using cross-entropy loss on these pairwise targets.

**Entropy-based continuous inference.** Although training was performed using pairwise classification, continuous predictions were obtained at inference using an entropy-based procedure. For each test sample, the trained model was evaluated against a set of reference samples with known target values drawn from the training set or a combination of the training and testing sets minus one image (whichever is currently being inferred). The resulting probabilities defined a target-dependent uncertainty profile, and the final prediction was taken as the reference value corresponding to maximal entropy.

**Computational cost.** Although pairwise construction yields $N(N-1)$ possible ordered comparisons from $N$ training samples, we do not enumerate or process all possible pairs during training. Instead, pairs are sampled and processed in mini-batches, keeping memory use bounded and training computationally practical. Our best UTKFace model reached its selected early-stopping checkpoint on a single RTX 2080 GPU in approximately 2.5 hours. At inference, a naive exhaustive pairwise procedure would require $O(C^2)$comparisons for $C$candidate/reference samples. The binary search tree (BST) reduces the number of candidate comparisons required for each prediction to $O(\log_2 C)$, yielding an overall inference complexity of $O(C\log_2 C)$rather than $O(C^2)$. For $C \approx 20{,}000$, this corresponds to approximately a 1,400-fold reduction in the number of comparisons. In practice, inference on 4,000 UTKFace test images, including reference comparisons, entropy-based decoding, BST traversal, and local refinement, required approximately 1.3 hours on the same GPU (about 1.2 s per test image).

## 2.2 UTKFace Age Regression

**UTKFace Dataset.** UTKFace is a publicly available face image dataset containing over 20,000 images annotated with age, sex, and ethnicity [32]. In this study, we used age labels only and treated age as a continuous prediction target within the proposed pairwise framework.

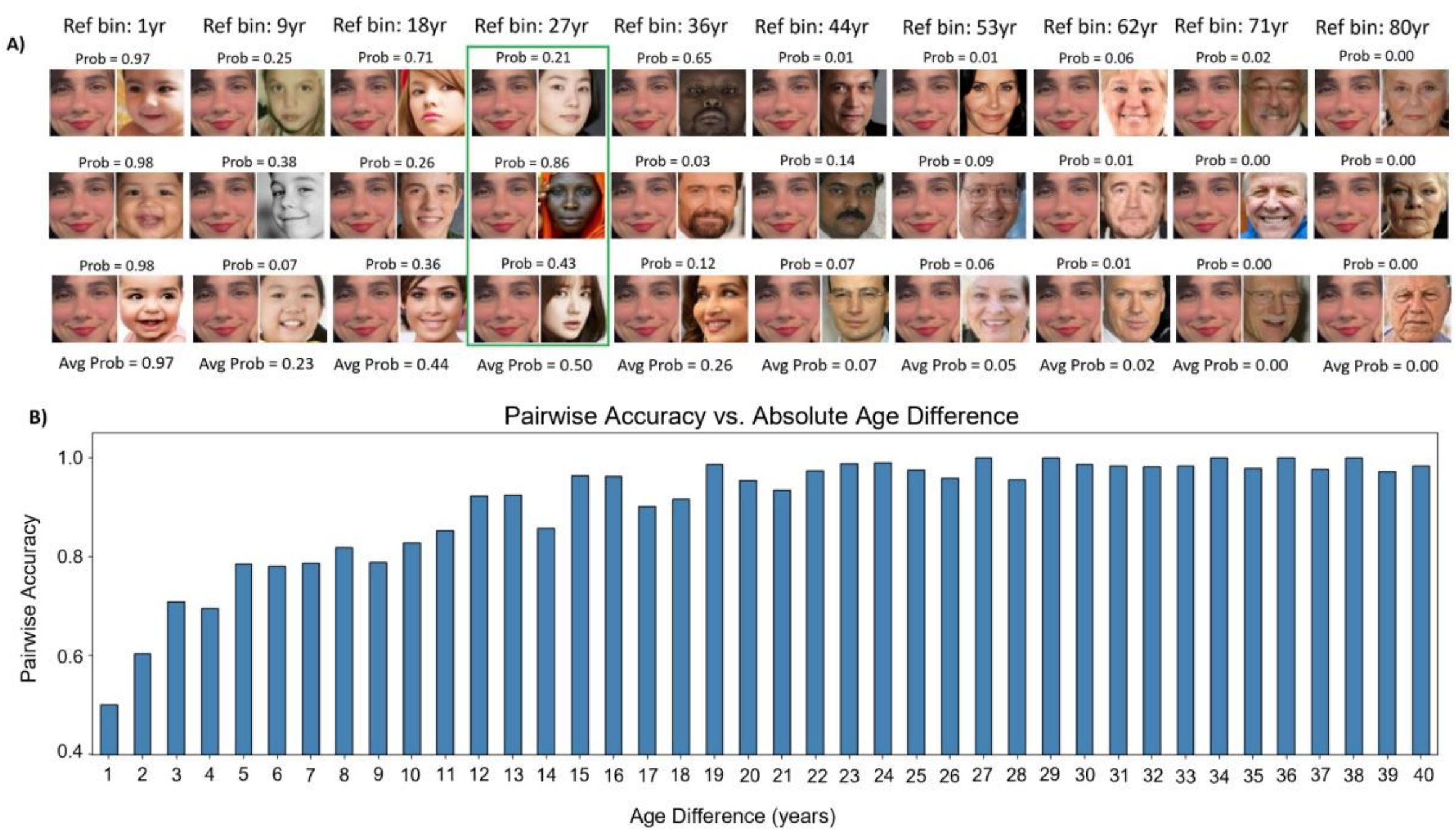

**Figure 2: Pairwise behavior of the age model on UTKFace. A)** Example grid of test and reference pairs used to visualize the model's internal reasoning. Each column corresponds to a reference age bin (not all bins shown), and each row shows several examples of test–reference pairs. The left image of each pair is the test face being compared against all known references, for this example, the model aggregates all comparisons and identifies the reference age bin with maximal entropy—here, the 27-year-old bin, where average "older" probability is closest to 0.5. **B)** Histogram showing pairwise accuracy as a function of absolute age difference $|y_i - y_j|$. The model is most uncertain when ages are similar, because cross-entropy imposes a logarithmic, unbounded penalty on confident incorrect predictions, motivating the entropy-based decoding strategy described in the text. The highest entropy is achieved when the 2 final layer's neurons output are equal (i.e. 50% probability). These two panels illustrate how pairwise supervision allows the network to infer a smooth relational manifold of aging rather than discrete labels.

**Reference self-consistency filtering.** It is known that UTKFace has tens of labeled data points that are either erroneous (e.g. the wrong age) or impossible to infer correctly (e.g. the face is just a picture of an eye and a little of the surrounding skin). We explored an optional procedure to clean the reference dataset of that low-quality data. Noisy or ambiguous reference data points were detected by inferring each one's age the same way test images are inferred (see Fig. 2). The error between the actual age and the inferred age is called self-consistency error. We discarded the 7.5% of references with the largest self-error. This filtering was used for Fig. 3A,C-D. Self-consistency filtering can be used to improve the performance on nearly any model at the cost of using more compute, but will be most beneficial when data is scarce or noisy.

**Pairwise decoding via entropy.** To visualize how the model reasons about age, we first examine a single test face and its pairwise comparisons against reference faces grouped into age bins. As illustrated in Fig. 2, the model's pairwise predictions vary smoothly with age difference: pairs with similar ages produce probabilities near 0.5 and high entropy, whereas pairs with large age gaps saturate toward 0 or 1.

**Setup.** UTKFace contains roughly 23k face images annotated with integer ages. We reserve approximately 4k images as a held-out test set and use the remainder for training and validation. During training, we form random pairs ($x_i$,$x_j$) within each mini-batch and supervise the model to predict which face is older. At inference time, to estimate the age of a test face $x_{\text{test}}$, we compare it against a set of reference faces in the training set (excluding $x_{\text{test}}$ itself), obtain pairwise "older/younger" probabilities across ages, and decode a scalar prediction via entropy-based search. We keep inference time quick and constant by applying a shallow binary search tree (BST) over the reference age bins, starting at 60, and then a local dense refinement over neighboring bins around the final tree node (e.g. +/- 5 years) (see Fig. 5B, for an example).

**Small scale test.** Regression collapse under small data regimes when training with pointwise losses serves as our main motivation for this work. Our first results are thus from our method applied to only a subset of 100 (face, age) pairs of the UTKFace dataset. Figure 1B shows how the error is nearly cut in half relative to the conventional direct pointwise MSE baseline in Fig. 1A, and the variety of outputs spans a much larger range instead of clustering around a single effective value in an attempt to minimize the loss. To determine whether the improvements observed in Figure 1 were robust across independent training runs, we repeated both the pairwise and pointwise models across five matched random seeds. Pairwise learning consistently reduced test MAE relative to the pointwise baseline for the UTKFace age prediction (paired t-test, $p$ = .003; Figure 1C). These results indicate that the improvement produced by pairwise learning is reproducible across model initializations rather than being driven by a single favorable training run. Next, we scale up the dataset size to see if this data-efficiency effect is enhanced for larger datasets.

**Effect of train/test/reference splits.** We next examine how performance depends on the amount of training data and the number of reference points used at inference. These effects are summarized in Fig. 3, which shows true versus predicted age for four representative UTKFace configurations (A–D).

With 4k training faces, 1k reference faces, and 1k test faces, the pairwise model already achieves a MAE of 5.93 years and makes sensible predictions across the full age range (Fig. 3A). Increasing both training and reference sets improves accuracy: when trained on roughly half of UTKFace and tested on the other half (12k/11k/4k), the model reaches 4.40 MAE on a large and challenging test set (Fig. 3B). Holding the training set fixed at 4k examples but increasing the number of reference points to 4,499 (4k/4,499/4k) reduces the MAE from 5.93 to 5.46 (Fig. 3C), illustrating that reference count alone can meaningfully improve performance even without additional training data. Because inference involves only forward passes with no gradient computation, there is no risk of test-set leakage during this phase, but we omit the current test image from the reference dataset just to be overly cautious and avoid further justification. Finally, Fig. 3D shows that the same pairwise formulation integrates seamlessly with a ResNet-18 pretrained on ImageNet, using only 1k UTKFace training images.

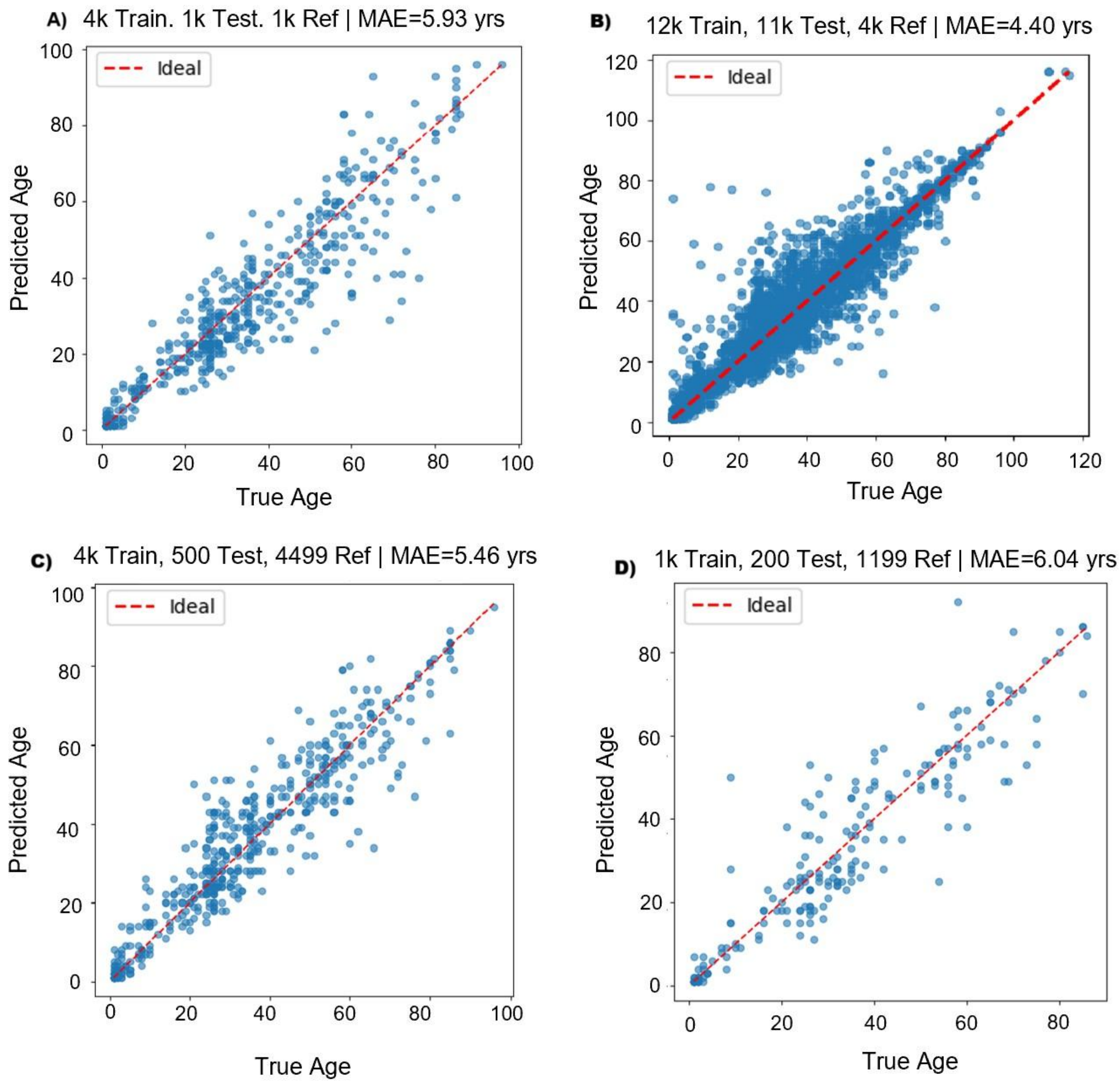

**Figure 3: Effect of training, reference, and backbone configurations on UTKFace age regression.** Pairwise regression affords state-of-the-art performance across the board, synergizing with nearly all deep learning methods and remaining extremely robust. **A)** As few as 4k training images and only 1k reference faces yield a robust predictor (MAE 5.93 yrs) that generalizes across the full age spectrum. **B)** Training on roughly half the dataset and testing on the other half (12k/11k/4k) produces a 4.40 MAE, demonstrating both scalability and generalization well beyond pretraining-based baselines. **C)** Holding the 4k training set fixed but increasing reference faces from 1k to 4,499 lowers MAE to 5.46, confirming that accuracy improves with richer reference sets even without retraining—since no learning occurs at inference, previously "test" data can be safely reused as references. **D)** A ResNet-18 pretrained on ImageNet achieves 6.04 MAE using only 1k training points, highlighting the flexibility of the pairwise framework and its seamless integration with transfer learning. A–C use scratch-trained ResNet-10s. Age ranges for each panel differ slightly depending on whether the full dataset (up to age 116, panel B) or a subset is used.

**Best configuration and SOTA performance.** Previous CNN-based approaches on UTKFace report MAEs ranging from approximately 4.6 to 9.7 years, depending on architecture and training strategy [27, 28]. Using the full training set, a scratch-trained ResNet-10 backbone, and the entropy+BST decoding described above, our best model achieves a mean absolute error of 3.64 years on the 4k-image test set (Fig. 4). This is numerically lower than the best (to our knowledge) previously reported UTKFace MAE of approximately 3.70 years, achieved using a transformer-based architecture [29], despite using a much smaller backbone, no pretraining, and only a single RTX 2080 GPU. We emphasize that this result uses neither reference filtering nor any model ensembling, it arises purely from pairwise supervision, simple augmentations, and a lightweight architecture.

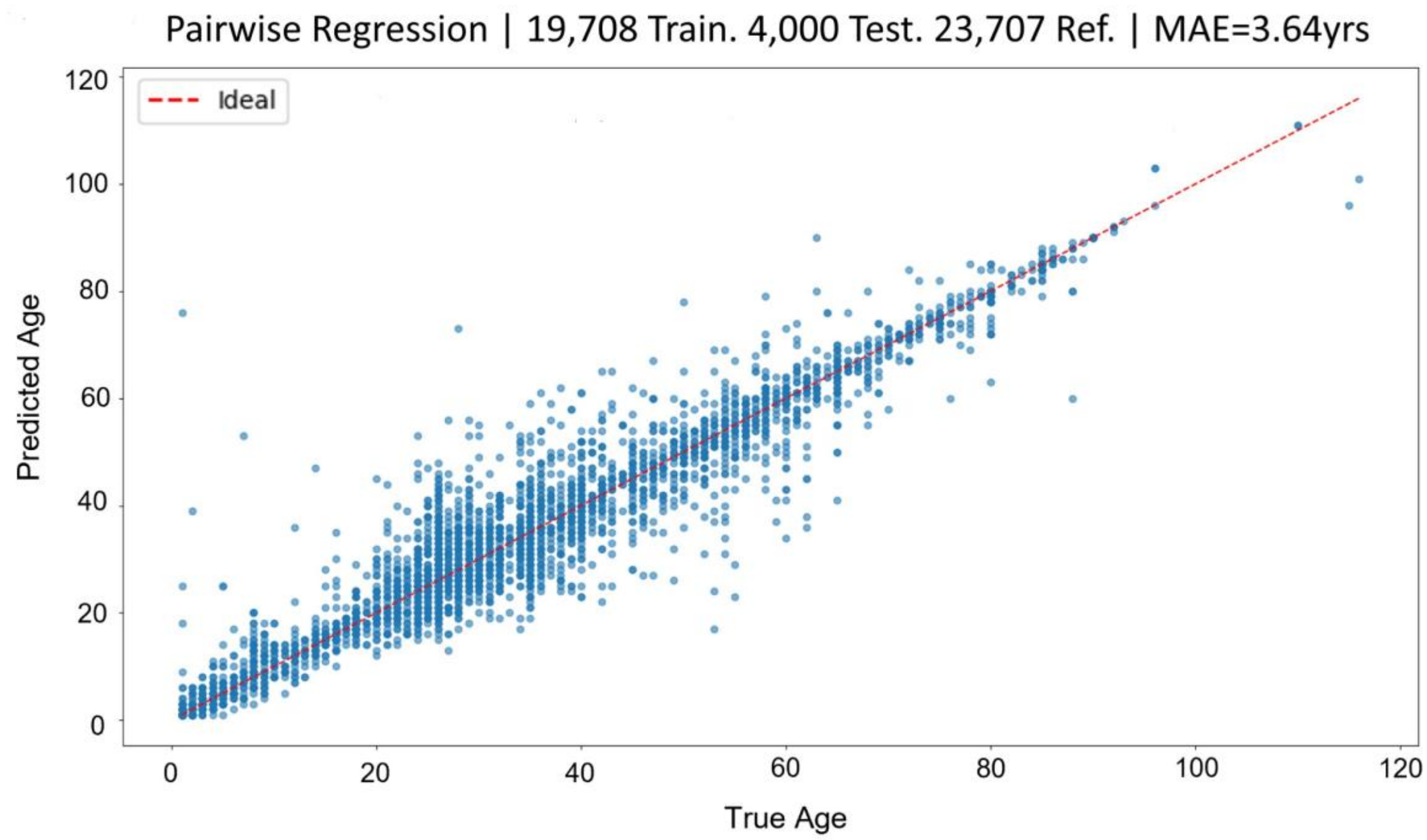

**Figure 4: Best-performing UTKFace model achieves state-of-the-art accuracy.** True versus predicted ages for our strongest configuration using 19,708 training, 4,000 testing, and 23,707 reference images. The model achieves a 3.64-year MAE, surpassing the previous best UTKFace benchmark (3.70 MAE) despite using a smaller scratch-trained ResNet-10, no pretraining, and only an RTX 2080 GPU. 8× test-time augmentation was enabled and no reference filtering was used.

**Learning dynamics and inference example.** Figure 5A shows training and test pairwise accuracy over epochs for the best UTKFace model. Training accuracy rises above 96%, while test accuracy plateaus around 91.4%, indicating that the network learns a stable ordering over ages without collapsing or catastrophically overfitting. Figure 5B illustrates a single BST + local dense inference trace: the model's predicted probability that the test face is older than each candidate age bin (blue curve) crosses 0.5 near the true age, and the corresponding entropy (red curve) peaks at the selected bin within the restricted local window.

**Transfer Learning with Pretrained ResNet-18** Although our primary focus is on lightweight models trained from scratch, it is natural to ask whether pairwise supervision can also leverage conventional pretrained backbones. To explore this, we conducted a transfer-learning experiment using ResNet-18 pretrained on ImageNet. We adapted the first convolutional layer to accept two-channel pairwise inputs and fine-tuned the network using a progressive unfreezing schedule: first training only the final linear head, then unfreezing the last residual block, and finally unfreezing the full model. The corresponding true-vs-predicted age scatter appears in Fig. 3D.

Using only 1,000 UTKFace training faces and a 200-image test set, the pretrained pairwise model achieves a mean absolute error of 6.04 years. During inference, each test face is compared against approximately 1,200 reference images, and its age is decoded using the same entropy-based aggregation as in the ResNet-10 experiments. This transfer-learning setup illustrates that pairwise supervision integrates naturally with pretrained networks and remains effective even when labeled data are extremely limited.

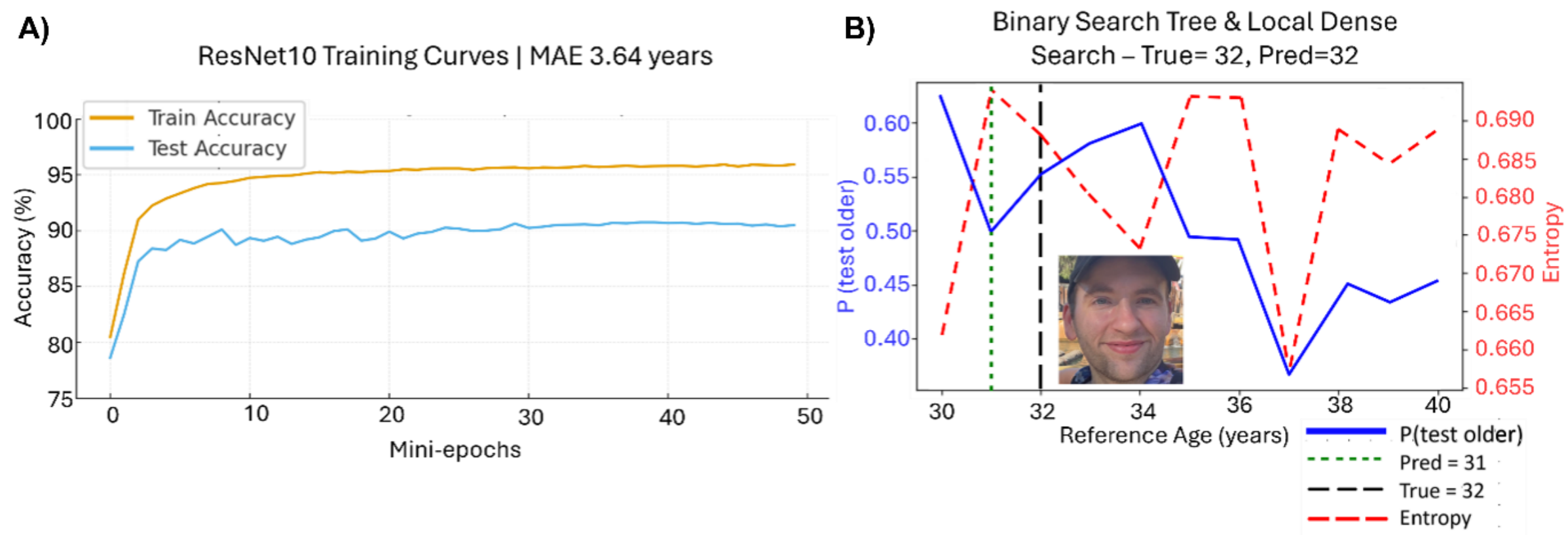

**Figure 5: Training dynamics and inference visualization. A)** Learning curves for the best UTKFace model. Training pairwise accuracy exceeds 96%, while test accuracy stabilizes around 91.4%, showing stable learning without overfitting. **B)** Example of binary search tree (BST) + local dense inference for a single non-UTKFace validation image. The blue curve shows the predicted probability that the test face is older than each reference age bin, the black dotted line marks the true age (32), and the green dotted line the predicted age. The red curve shows entropy, peaking near 0.5 probability (blue), which is where it makes the age prediction (green), the BST after three rounds converged on the 35-year node and restricted the search to a ±5-year window (30–40), where the local dense search produced the final estimate.

**Pointwise MSE baselines** For a small subset of UTKFace we compare against pointwise baseline trained with a standard loss using the same backbone with minimal modifications (see Fig. 1). For these we tested several output-normalization strategies. Bounding the regression head with a tanh nonlinearity frequently caused saturation, while leaving the output unnormalized yielded unstable gradients. A simple linear rescaling of the final output produced the most reliable learning dynamics and was adopted for all MSE results.

## 2.3 STL-10 Classification

**STL-10 Dataset.** STL-10 is a publicly available natural image dataset consisting of over 100,000 96×96 color images from 10 object categories (airplane, bird, car, cat, deer, dog, horse, monkey, ship, and truck) [31]. We used the labeled subset and constructed image pairs for a binary classification task indicating whether two images belonged to the same semantic class.

*Code Availability***.** All code is publicly available at https://github.com/bainro/pairwise_learning.

**Setup.** To test whether pairwise supervision also benefits classification, we consider STL-10, which provides 5,000 labeled images across 10 classes of human-made objects and is commonly used for low-data experiments [31]. We compare two models with identical convolutional backbones: (i) a standard classifier that takes single images and is trained with a 10-way cross-entropy loss, and (ii) a pairwise model that takes image pairs and is trained only to decide whether the pair of images belong to the same class or not. The pairwise model never sees direct single-image labels during training. To probe the low-data regime, we train both models on subsets of 100, 500, 1,000, and 2,000 labeled images, keeping the test set fixed.

**Results and visualization.** Across the different training set sizes, pairwise supervision outperformed the pointwise baseline at 500, 1,000, and 2,000 training images, with the largest mean gain observed at 2,000 images. The advantage across multiple training-set sizes indicates that the benefit is not tied to a single operating point. Figure 6B visualizes the class-wise output distribution for a correctly classified truck image. Across five independent runs, pairwise supervision achieved significantly higher test accuracy than the baseline for training subsets of 500 ($t(4)=8.99$, $p<0.001$), 1,000 ($t(4)=3.95$, $p=0.008$), and 2,000 images ($t(4)=10.90$, $p<0.001$), whereas no significant difference was observed at 100 images ($p=0.905$). Averaged across all training subset sizes, pairwise supervision also yielded significantly higher accuracy than the baseline ($t(4)=9.18$, $p<0.001$).

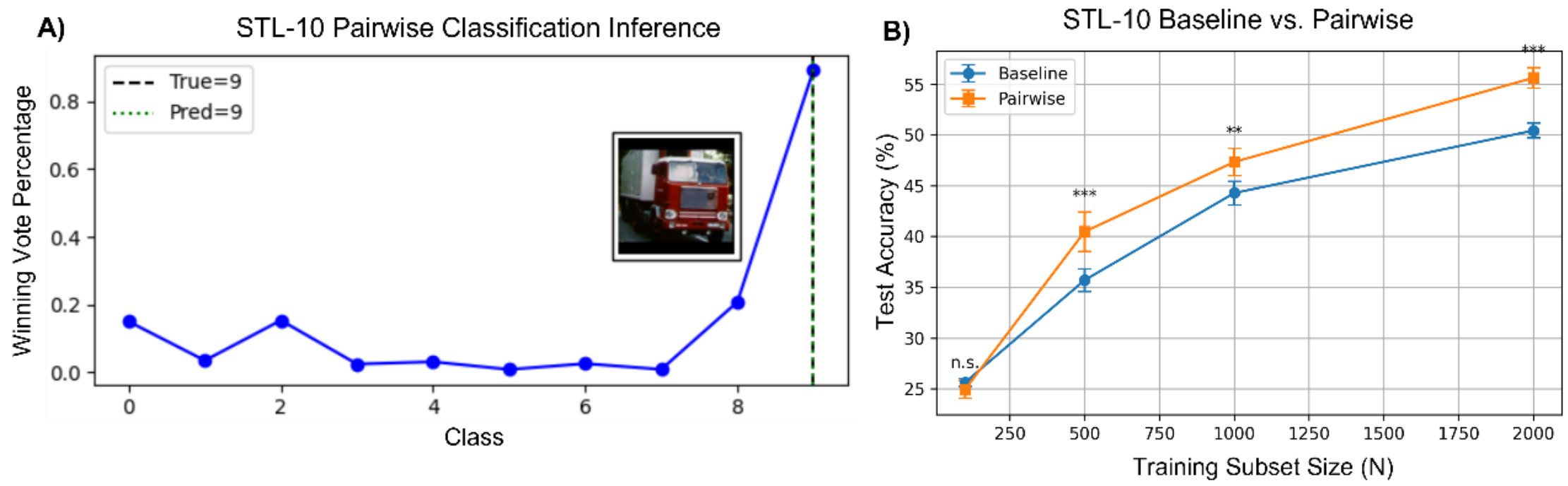

**Figure 6: Pairwise supervision improves STL-10 classification in low-data regimes. A)** Class-wise output distribution for an example STL-10 test image (a truck). The model correctly assigns maximal probability to class 9, with human-made object classes clustering separately from animal categories. **B)** Test accuracy of pairwise (red) versus standard cross-entropy classifiers (blue) across training subsets of 100, 500, 1,000, and 2,000 labeled samples. Points show mean accuracy across five independent runs, with error bars indicating standard deviation. Pairwise supervision achieved significantly higher accuracy at 500 ($p<0.001$), 1,000 ($p=0.008$), and 2,000 ($p<0.001$) training samples, whereas no significant difference was observed at 100 samples.

# 3 Discussion and Conclusion

Deep neural networks are notoriously fragile in small-sample regression, often collapsing toward low variance predictions that simply reproduce the training-set mean. In this work we showed that reframing regression as pairwise comparison—learning to decide which of two inputs should receive the larger target value—can dramatically improve data efficiency without changing the underlying architectures. Across two very different settings (face age estimation and low-data image classification), the same simple pairwise framework reliably outperformed pointwise baselines trained with standard losses.

On UTKFace, pairwise supervision turned a modest scratch-trained ResNet-10 [30] into a state-of-the-art age regressor, achieving 3.64-year MAE on a 4,000-image test set using a single RTX 2080 GPU and no pretraining or ensembling (Fig. 4). The method gains much of its power from the implicit $O(N^2)$ constraint graph over the data: every pair of faces contributes a local ordering, and the model learns a smooth age manifold that behaves sensibly even where labeled examples are sparse. The entropy-based decoding plus binary search tree (BST) refinement further makes inference efficient and interpretable, as visualized in our pairwise probability and entropy plots (Figs. 2 and 5). Importantly, the same pairwise formulation worked across a range of train/reference/test splits and integrated naturally with a pretrained ResNet-18 (Fig. 3D), suggesting that the benefits are robust to both data regime and backbone choice.

On STL-10, pairwise training also improved supervised classification in the low-data regime. With identical backbones, the pairwise model consistently outperformed the standard 10-way cross-entropy classifier across three out of four training-set sizes (500, 1,000, and 2,000 labeled examples), with an average gain of 5.3 percent and a peak improvement of 8.3 percent at 500 labels (Fig. 6). Intuitively, pairwise labels are more informative than arbitrary class IDs: rather than only learning that an image belongs to class $k$, the network learns a relative ordering over classes and images, leading to softer, more graded signals about which examples should be separated in representation space. This richer supervision appears to induce more robust features and better class separation when labeled data are scarce.

Conceptually, our framework is related to prior pairwise learning approaches that replace pointwise supervision with relationships between examples. Learning-to-rank methods such as RankNet and related preference-learning approaches [22, 39] likewise optimize pairwise ordering constraints rather than absolute predictions. Our classification formulation is also related to Siamese and metric-learning approaches, which learn from pairs labeled according to similarity or class membership [35, 36]. However, our goal differs from these approaches: rather than primarily learning a ranking function or embedding space, we use pairwise supervision as a general replacement for standard pointwise regression and classification losses and then recover the original prediction task at inference. Moreover, whereas some relative-learning approaches obtain explicitly annotated comparisons or relative attributes [37, 38], our pairwise targets are generated automatically from the original scalar or categorical labels and therefore require no additional annotation.

Beyond these conceptual connections, our results provide some insight into why this formulation may be particularly useful for small-data prediction. In the UTKFace regression experiments, the most striking difference from pointwise learning was not simply lower error, but preservation of the target's dynamic range: the pointwise models tended to collapse toward the training-set mean, whereas pairwise supervision produced predictions that varied across much more of the observed target range. This suggests that the advantage may arise from replacing the difficult task of directly calibrating an absolute continuous value from each individual example with the simpler local question of which of two examples has the larger target, while recovering the absolute value only at inference through comparisons with labeled references. In this respect, our results extend the intuition behind learning-to-rank and preference-learning methods: relative supervision can be useful not only for producing rankings, but also as an intermediate representation from which continuous values can be reconstructed. The STL-10 results provide a complementary observation. Pairwise supervision significantly improved classification at 500, 1,000, and 2,000 training examples, but not at 100, suggesting that pairwise construction does not automatically overcome arbitrarily severe data scarcity, rather, it appears to become advantageous once there are enough labeled examples to learn meaningful relationships among samples. Together, these findings suggest that the benefit of pairwise supervision comes from restructuring the prediction problem around relationships that are easier to learn from limited data, while retaining the ability to recover the original regression or classification target.

A key practical advantage of this framework is that it does not require large models or massive compute. The core change is entirely in the loss and label structure. For regression, if $y_i > y_j$, the pair $(x_i, x_j)$ provides one ordering constraint and the reversed pair $(x_j, x_i)$ provides the opposite constraint. Thus, each labeled example can participate in many supervised pairwise relationships without additional annotation. For classification, the same principle allows pairs to be labeled as belonging to the same or different classes using only the original class labels. In our experiments, even a small ResNet-10 trained from scratch was able to achieve pairwise accuracy above 96% on UTKFace (Fig. 5), yet remained cheap enough to train in a couple of hours on a single consumer GPU. The heavy lifting comes from recycling existing labels across many pairs, not from scaling up model size or training duration.

Despite these advantages, the approach is not without drawbacks. First, the number of possible training pairs grows quadratically with the dataset size. Although we never enumerate all pairs in practice—and our best UTKFace result remained tractable on a single GPU—very large datasets will still require careful subsampling or curriculum strategies to keep training efficient. Second, performance depends on the quality and coverage of the reference set used at inference: if references are heavily biased or missing important regions of the target space, the decoded scalar values will inherit those biases. Third, decoding adds an extra layer of complexity compared to standard regression: entropy-based search, BST routing, and local dense refinement are simple, but still more involved than a single forward pass through a pointwise regressor. Finally, while our visualizations provide some intuition, we have not yet provided a formal theoretical characterization of when pairwise supervision is guaranteed to outperform pointwise losses, or how it interacts with label noise and systematic annotation biases.

There are several natural extensions of this work. Methodologically, pairwise supervision could be combined with more sophisticated ranking or listwise objectives, uncertainty-aware decoding, or active selection of informative pairs, potentially improving sample efficiency and robustness. For classification, pairwise training provides a principled framework for ideas often used heuristically, such as soft-label generation, sample biasing, and reference-set balancing, which may yield better calibrated models under class imbalance or label noise. Architecturally, it would be interesting to test the framework with larger pretrained vision transformers and multimodal models, examining whether pairwise supervision can extract more value from limited domain-specific labels than standard fine-tuning. Finally, the STL-10 experiment suggests the approach may generalize to other small-sample, high-stakes domains such as clinical populations, histopathology, or behavioral scoring from video.

This research did not receive any specific grant from funding agencies in the public, commercial, or not-for-profit sectors.